\documentclass{cimart}

\usepackage{booktabs}
\usepackage{gensymb}
\usepackage{subcaption}
\usepackage{tabularx}
\usepackage{textcomp}
\usepackage{threeparttable}

\title{A digital pheromone-based approach for in-control/out-of-control classification}

\authors{Pedro Pestana and Maria de F\'atima Brilhante}

\authorinfo[P. Pestana]{Aberta University, Portugal}{pedro.pestana@uab.pt}

\authorinfo[M. F. Brilhante]{University of the Azores, Portugal}{maria.fa.brilhante@uac.pt}

\abstract{In complex production lines, it is essential to have strict, fast-acting rules to determine whether the system is In Control (InC) or Out of Control (OutC). This study explores a bio-inspired method that digitally mimics ant colony behavior to classify InC/OutC states, enabling the detection of process transitions requiring maintenance. A case study on industrial potato chip frying provides the application context. During each two-minute frying cycle, sequences of eight temperature readings are collected. Each sequence is treated as a digital ``ant'' depositing virtual pheromones, generating a Base Score. New sequences, representing new ants, can either reinforce or weaken this score, leading to a Modified Base Score that reflects the system’s evolving condition. Signals such as extreme temperatures, large variations within a sequence, or the detection of change-points contribute to a Threat Score, which is added to the Modified Base Score. Since pheromones naturally decay over time unless reinforced, an Environmental Score is incorporated to reflect recent system dynamics, imitating real ant behavior. This score is calculated from the Modified Base Scores collected over the past hour. The resulting Total Score, obtained as the sum of the Modified Base Score, Threat Score, and Environmental Score, is used as the main indicator for real-time system classification and forecasting of transitions from InC to OutC. This ant colony optimization-inspired approach provides an adaptive and interpretable framework for process monitoring and predictive maintenance in industrial environments.}


\acknowledgments{The authors are grateful for the permission to use and archive the 2025 data they were asked to analyze, provided that the company’s trademark remains undisclosed to protect its reputation. This work is funded by national funds through FCT --- Funda\c{c}\~ao para a Ci\^encia e a Tecnologia, I.P., under CIAC Research Unit, UID/04019/2025, DOI: \url{https://doi.org/10.54499/UID/04019/2025}, and under CEAUL Research Unit, UID/00006/2025, DOI: \url{https://doi.org/10.54499/UID/00006/2025}. The authors declare no conflict of interest, and have contributed equally to this work.}


\VOLUME{34}
\YEAR{2026}
\ISSUE{1}
\NUMBER{7}
\DOI{https://doi.org/10.46298/cm.16690}
\editinfo{October 10, 2025}{February 20, 2026}{Ana Cristina Moreira Freitas}

\begin{document}

\section{Introduction}\label{Intro}
 
 In a production line, the system is expected to remain In Control (InC) under normal operating conditions. However, a wide range of factors can lead to an Out of Control (OutC) state, often requiring production interruptions for maintenance or repair. Both false discoveries (i.e., incorrectly identifying the system as OutC) and false omissions (i.e., failing to detect a true OutC state) can result in significant financial losses. Therefore, prompt and accurate classification of the system as InC or OutC, along with the ability to predict imminent transitions from InC to OutC, is of critical importance in Statistical Quality Control (SQC). To ensure reliability, such classification systems must be evaluated not only for their speed and accuracy but also in terms of their sensitivity and specificity. 

The authors were previously consulted to conduct a detailed analysis of a fried chip production line, using 2024 factory data converted to a nominal scale, with a focus on oil temperature measurements in the frying unit. 
The analysis accounted for sensitivity and specificity in diagnosing InC and OutC states. Under InC conditions, the target oil temperature is 180~\degree\text{C}. However, the introduction of new batches of raw potato slices inevitably causes a drop in temperature, while fluctuations due to Joule heating may induce temperature increases. Temperatures consistently below 170~\degree\text{C} negatively affect crisping, de-oiling, and seasoning, whereas temperatures persistently above 190~\degree\text{C} may lead to starch burning, compromising both texture and flavor.

Based on the recommendations the authors made regarding data collection protocols, record-keeping, and the implementation of improved feedback and control policies, significant operational enhancements were observed. From January 2025 onward, temperature data began to be recorded on a ratio scale (see Stevens~\cite{Stevens}), allowing for more precise and responsive decision-making processes. In particular, digital pheromones, inspired by Ant Colony (AC) behavior and the pioneering work on Ant Colony Optimization (ACO) by Dorigo~\cite{Dorigo1992}, were adopted to score sequences of temperature readings collected during each 2-minute frying cycle. For comprehensive overviews of AC behavior and its applications, see Dorigo and St\"utzle~\cite{AntCol}, Monmarch\'e et al. \cite{ArtAnt}, and Martens et al. \cite{Martens2011}.
 
Classification algorithms based on ACO have been developed by Parpinelli et al. \cite{Parpinelli}, Martens et al. \cite{Martens}, and Otero et al. \cite{Otero}.  Zhang et al. \cite{Zhang} proposed an enhanced genetic AC algorithm for bankruptcy prediction, which is relevant to our forecasting context. The quality of the classification results is evaluated using confusion and matching matrices, which capture the agreement and non-agreement between predicted classifications, i.e., Diagnosed InC (DInC) and Diagnosed OutC (DOutC), and the actual outcomes, namely, InC (Negative) and OutC (Positive).

In addition to the total numbers of Negatives (N) and Positives (P), classification performance is also described in terms of the confusion-matrix outcomes: True Negatives (TN), False Positives (FP), True Positives (TP), and False Negatives (FN). To facilitate the interpretation of these quantities, indicated in Table~\ref{confus2}, a conditional-probability-style notation can be adopted:
\begin{itemize} 
 \item True Negatives: $\mathrm{TN=DInC \vert InC}$ --- correctly identifying an in-control state;
\item False Positives: $\mathrm{FP=DOutC \vert InC}$ --- diagnosing the system as out-of-control when it is actually in control;
 \item True Positives: $ \mathrm{TP=DOutC \vert OutC}$  --- correctly identifying an out-of-control state; 
 \item False Negatives: $\mathrm{FN =DInC \vert OutC}$ --- diagnosing the system as in-control when it is actually out of control.
  \end{itemize}
  
Other performance metrics presented in Table~\ref{confus2} are defined in Table~\ref{param}. 
For a comprehensive discussion of confusion matrices and associated metrics, the reader is referred to \cite{Stehman, Powers, TingE, TingCM, TingSS, Tharwat, Chicco, Chicco2, Opitz}.

\begin{table}[htbp]
\centering
\caption{Confusion matrix}\label{confus2}
\begin{tabular}{|l|c|c|c|c|} \hline
Total&\bf{DOutC} & \bf{DInC}& BM  &PT  \\ \hline
\bf{OutC (P)} & \bf{TP}  & \bf{FN}  & TPR  & FNR \\ \hline
\bf{InC (N)} & \bf{FP}  & \bf{TN}  & FPR  & TNR  \\ 
\hline
Prev  & PPV  & NPV  & LR${}^+$   & LR${}^-$  \\ \hline
Acc & FDR   & FOR  & MK $\Delta_P$ & DOR  \\ \hline
BAcc   & $\text{F}_1$ Score   & FM  Index & MCC  & CSI  \\ \hline
\end{tabular}
\end{table}

  \begin{table}[htbp]
\renewcommand{\arraystretch}{1.2}
\centering
\caption{Performance metrics}\label{param}
\begin{threeparttable}
\begin{tabular}{ll}
\toprule
Metric & Abbreviation and definition \\
\midrule
Sensitivity or True Positive Rate & $\rm{TPR=\frac{TP}{P}}$\\
Specificity or True Negative Rate & $\rm{TNR=\frac{TN}{N}}$\\
False Negative Rate & $\rm{FNR=\frac{FN}{P}}$\\
False Positive Rate & $\rm{FPR=\frac{FP}{N}}$\\
Prevalence & $\rm{Prev=\frac{P}{P+N}}$\\
Positive Predictive Value (Precision) & $\rm{PPV=\frac{TP}{TP+FP}}$\\
Negative Predictive Value & $\rm{NPV=\frac{TN}{TN+FN}}$\\
False Discovery Rate & $\rm{FDR=\frac{FP}{TP+FP}}$\\
False Omission Rate & $\rm{FOR=\frac{FN}{TN+FN}}$\\
Positive Likelihood Rate & $\rm{LR^{+}=\frac{TPR}{FPR}}$\\
Negative Likelihood Rate & $\rm{LR^{-}=\frac{FNR}{TNR}}$\\
Accuracy & $\rm{Acc=\frac{TP+TN}{P+N}}$\\
Balanced Accuracy & $\rm{BAcc=\frac{TPR+TNR}{2}}$\\
Bookmaker Informedness & $\rm{BM=TPR+TNR-1}$\\
Markedness & MK $\rm{\Delta_P=PPV+NPV-1}$\\
Prevalence Threshold & $\rm{PT=\frac{\sqrt{TPR\times FPR}-FPR}{TPR-FPR}}$\\
Diagnosis Odds Ratio & $\rm{DOR=\frac{LR^{+}}{LR^{-}}}$\\
${\rm{F_1}~Score}$ & $\rm{F_1\,Score=\frac{2PPV\times TPR}{PPV+TPR}=\frac{2TP}{2TP+FP+FN}}$\\
Fowlkes-Mallows Index & $\rm{FM=\sqrt{PPV\times TPR}}$\\
Critical Success Index CSI\tnote{(1)} & CSI=$\rm{\frac{TP}{TP+FN+FP}}$\\
Matthews Correlation Coefficient &
  $\rm{MCC=\sqrt{TPR\times TNR\times PPV\times NPV}}$\\
  & \qquad\qquad$\rm{-\,\sqrt{FNR\times FPR\times FOR\times FDR}}$\\
\bottomrule
\end{tabular}
\begin{tablenotes}
\footnotesize
\item[(1)] Or Jaccard Index JI.
\end{tablenotes}
\end{threeparttable}
\end{table}

In this paper, Section~\ref{Data} describes the data, which consists of temperature readings collected during potato chip production. 
Section~\ref{SQC} demonstrates that classical SQC methods are not adequate for the data analyzed, thereby motivating the development of a more robust approach for defining reliable stopping rules.
Section~\ref{Method} outlines the new methodology used to score temperature sequences. The scoring framework, comprising base scores, modified base scores, environmental scores, threat scores, and total scores, is inspired by the Common Vulnerability Scoring System (CVSS) \cite{CVSS}. These scores are interpreted as digital pheromones contributing to the classification between DInC and DOutC.

In Section \ref{sec4}, the classification performance is evaluated. This evaluation is based on false discoveries and false omissions measures derived from the confusion and matching matrices.
Section \ref{Conclusions} summarizes the main findings, with a comparison to the authors' earlier results, which relied on less structured internal data from 2024; it also discusses the observed limitations. These reflections bring to mind Sir Ronald Fisher’s well-known aphorism at the First Session of the Indian Statistical Conference in Calcutta (1938): ``\textit{To consult the statistician after an experiment is finished is often merely to ask him to conduct a post-mortem examination.}''

\section{The data}\label{Data}
   
The analyzed data consist of oil temperature measurements collected during potato chip production from January 2 to April 30, 2025. Production runs Monday through Saturday, operating continuously from 07:00 a.m. until 03:00 a.m. the following day.
 During each 2-minute frying session, 8 temperature readings are taken at 15-second intervals. The system includes several feedback and control mechanisms designed to maintain operation within the InC state, defined by a Gaussian distribution with a mean of  \mbox{$\mu=180\degree$} and a standard deviation of \mbox{$\sigma=4\degree$} (the Celsius symbol is omitted throughout for simplicity). Large deviations, i.e., \mbox{$\pm 2.5\,\sigma = \pm 10\degree$}, can negatively impact product quality: lower temperatures may impair crispiness and disrupt downstream processes such as de-oiling and seasoning, while excessively high temperatures risk burning the starch, leading to undesirable taste and flavor profiles.

On the other hand, the rheostat regulating the Joule heating effect can sometimes introduce excessive corrections, namely, by cooling the oil too much after temperature spikes or overheating after persistently low temperature sequences, such as when all readings fall below 180\degree. Since production supervisors are not particularly inclined to work directly with numerical data, historical practices included the use of visual indicators  based on an ordinal scale inspired by Stevens~\cite{Stevens}. Within the framework of increasing temperatures, this system used color-coded annotations to represent sequences: 
\begin{itemize}
\item Cells with all temperatures between 184\degree\ and 188\degree\ were marked in orange;
\item Temperatures exceeding 188\degree\ were marked in red;
\item Sequences with all values below 180\degree\ were filled in blue;
\item If the initial temperature was below 174\degree, the cell was marked violet.
\end{itemize}

These color classifications were grounded in the properties of the Gaussian distribution. That is, the thresholds were chosen to correspond to typical deviations from the mean in the InC state, allowing for intuitive visual cues: orange indicates values within one standard deviation from the mean, red signals unusually high values (greater than two standard deviations), blue highlights consistently below-average conditions, and violet flags extremely low initial values, which is often indicative of system anomalies or start-up instability.

In addition, if a change in the mean of the distribution was detected, indicating a potential transition from InC to OutC, the sequence was labeled with a change-point (CP) tag. Multiple change-points could be identified within a single sequence. While various change-point detection methods are available (see Truong et al. \cite{Truong} and Burg and Williams~\cite{Burg} for reviews), we employed the binary segmentation algorithm (Scott and Knott~\cite{Scott}; Sen and Srivastava~\cite{Sen}). The Pruned Exact Linear Time (PELT) algorithm (Killick et al. \cite{killick}) was also initially considered for the task, using the Akaike Information Criterion (AIC) as the penalty term. However, it occasionally identified 4 to 5 change-points within a single sequence,  an outcome considered excessive given the length of the sequences. In contrast, the binary segmentation algorithm detected no more than 3 change-points per sequence. Since the final scores obtained using both algorithms were only marginally different, binary segmentation was considered for change-point detection.

\begin{center}
\begin{figure}[htbp]
\centering
\includegraphics[scale=0.6]{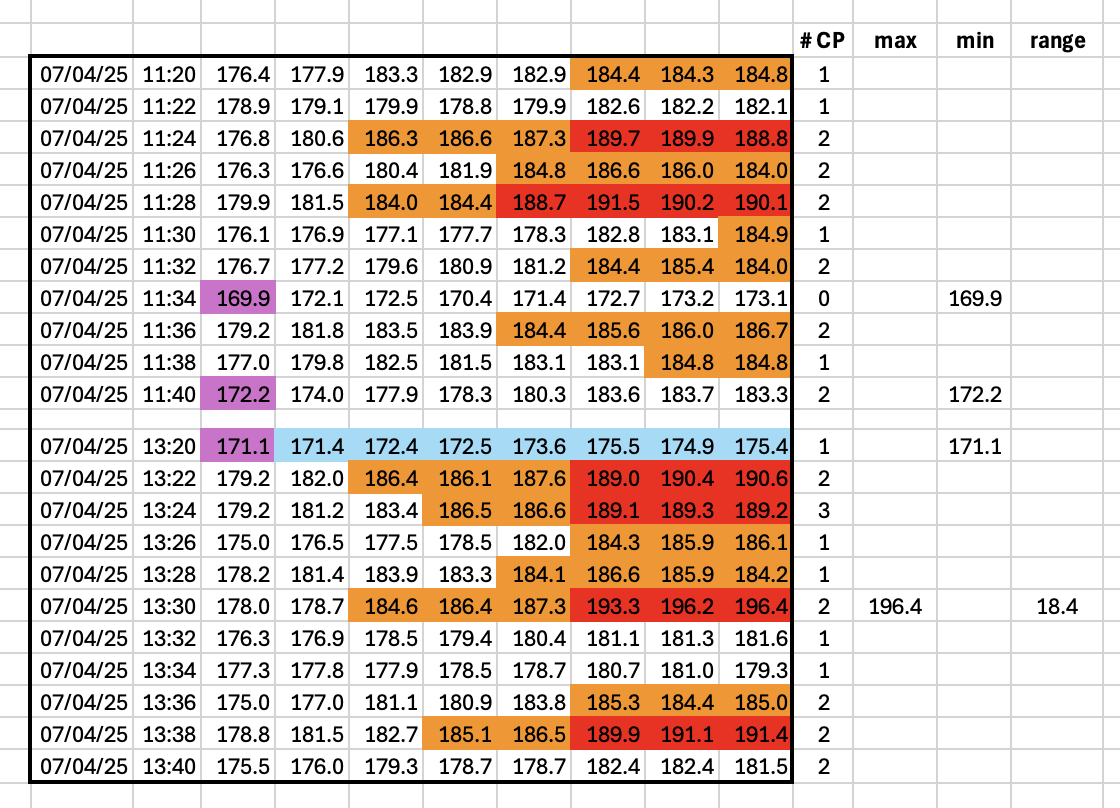}
\caption{Data extracted from production on April 7}\label{diar}
\end{figure} 
\end{center}

 Sequences were also annotated for extreme values, defined as maxima exceeding 195\degree, minima below 174\degree, or a  range greater than 13\degree. These annotations are displayed alongside each sequence, as illustrated in Figure~\ref{diar}, which shows a typical data visualization as seen on the supervisors' monitor screens.

\section{Shortcomings of classical SQC methods}\label{SQC}

SQC is a well-established discipline within Statistics. In an ideal setting, the application of methodologies described in good textbooks, such as Montgomery \cite{Montgomery}, together with the use of robust analytical software, would allow for a rigorous and balanced diagnosis of weaknesses in factory data analysis and the formulation of appropriate and effective corrective actions. In practice, however, the realities of data collection fall far short of this ideal. As De Veaux and Hand \cite{Hand} emphatically observe in the opening line of their paper’s introduction, ``\textit{Bad data can ruin any analysis.}'' They further note that ``\textit{Anyone who has analyzed real data knows that the majority of their time on a data analysis project will be spent `cleaning' the data before doing any analysis.}'' We show that, in this setting, classical methodologies fail not because the methodologies themselves are flawed, but because of the presence of messy data.

\begin{figure}[htbp]
\centering
\includegraphics[scale=0.17]{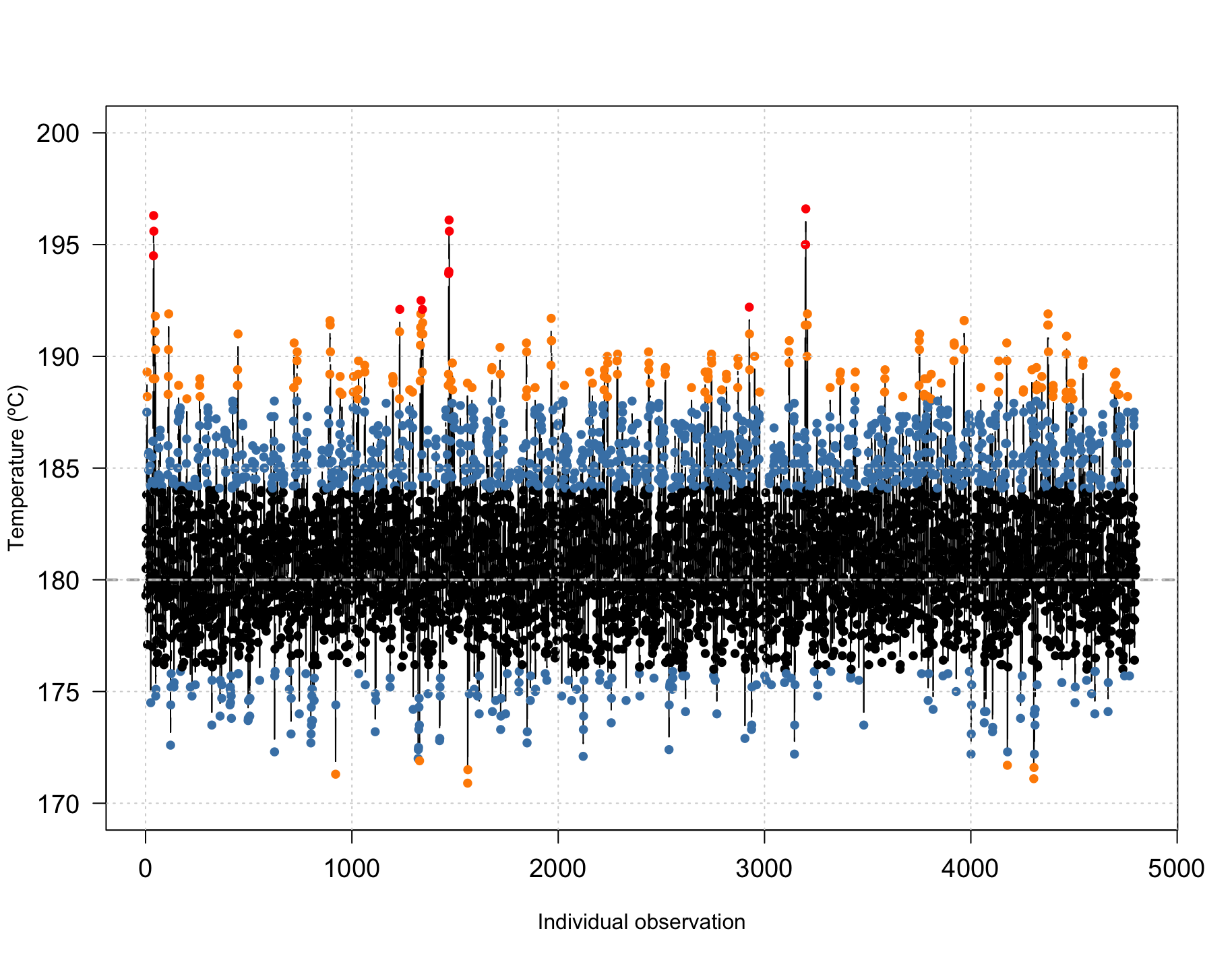}
\caption{Daily temperature time series for January 2. Points are color-coded by deviation from the target mean 
\mbox{$\mu=180\degree$}: black points fall within $\pm 1\sigma$, blue points are between $\pm 1\sigma$ and $\pm 2\sigma$, orange points are between $\pm 2\sigma$ and $\pm 3\sigma$, and red points exceed $\pm 3\sigma$.}\label{Jan02_ts}
\end{figure}

Figure~\ref{Jan02_ts} displays the daily temperature readings (\mbox{$n=4{,}800$}) collected on January~2, a day on which no production halt occurred. Approximately 72\% of the recorded temperatures fall within the nominal $1\sigma$ limits ($[176\degree, 184\degree]$), while 0.29\%  fall outside the nominal $3\sigma$ limits ($[168\degree, 192\degree]$). These statistics reflect the observed distribution of measurements and may deviate from the theoretical expectations under a Gaussian assumption. Moreover, due to the continuous nature of the frying process, successive temperature observations exhibit temporal dependence, resulting in inherent autocorrelation (lag-1 autocorrelation $=0.634$).

Classical SQC methods for variables assume independence among subgroups. In this case, temperature data were collected in sequences of 8 observations, recorded every 15 seconds over 2-minute windows, with each sequence corresponding to the frying period for a batch of potato chips and used by the factory to classify the process as  InC or OutC. By treating each sequence of 8 temperature measurements as a rational subgroup and monitoring the corresponding subgroup mean, the effect of autocorrelation between consecutive sequences is reduced through temporal aggregation. Consequently, when the process is InC, the resulting sequence of subgroup means can be regarded as approximately independent, thereby justifying the use of classical control charts to assess  the process status, even in the presence of residual thermal inertia between successive sequences. Furthermore, the normality of the subgroup mean is ensured.

The Shewhart $\bar{x}$ chart is well suited to monitoring subgroup means and is particularly effective at detecting moderate to large shifts in the process mean, typically on the order of $2\sigma$ or more (see Montgomery \cite{Montgomery}). This matches the factory’s objective of minimizing unnecessary process stoppages and the associated production costs. By contrast, an Exponentially Weighted Moving Average (EWMA) chart would be more suitable if the primary goal were to detect smaller shifts in the process mean.
 Although time series modeling could account for the autocorrelation in the raw data, it was not pursued here because the factory’s priority is to preserve simple, actionable rules for stopping the process when it is OutC. In the present context, however, classical control charts are not adequate, as they would prompt an unacceptably large number of stoppages, far more than those observed under the factory’s current stopping rules.

\begin{figure}[htbp]
\centering
\begin{minipage}{0.48\textwidth}
\centering
  \includegraphics[width=\linewidth]{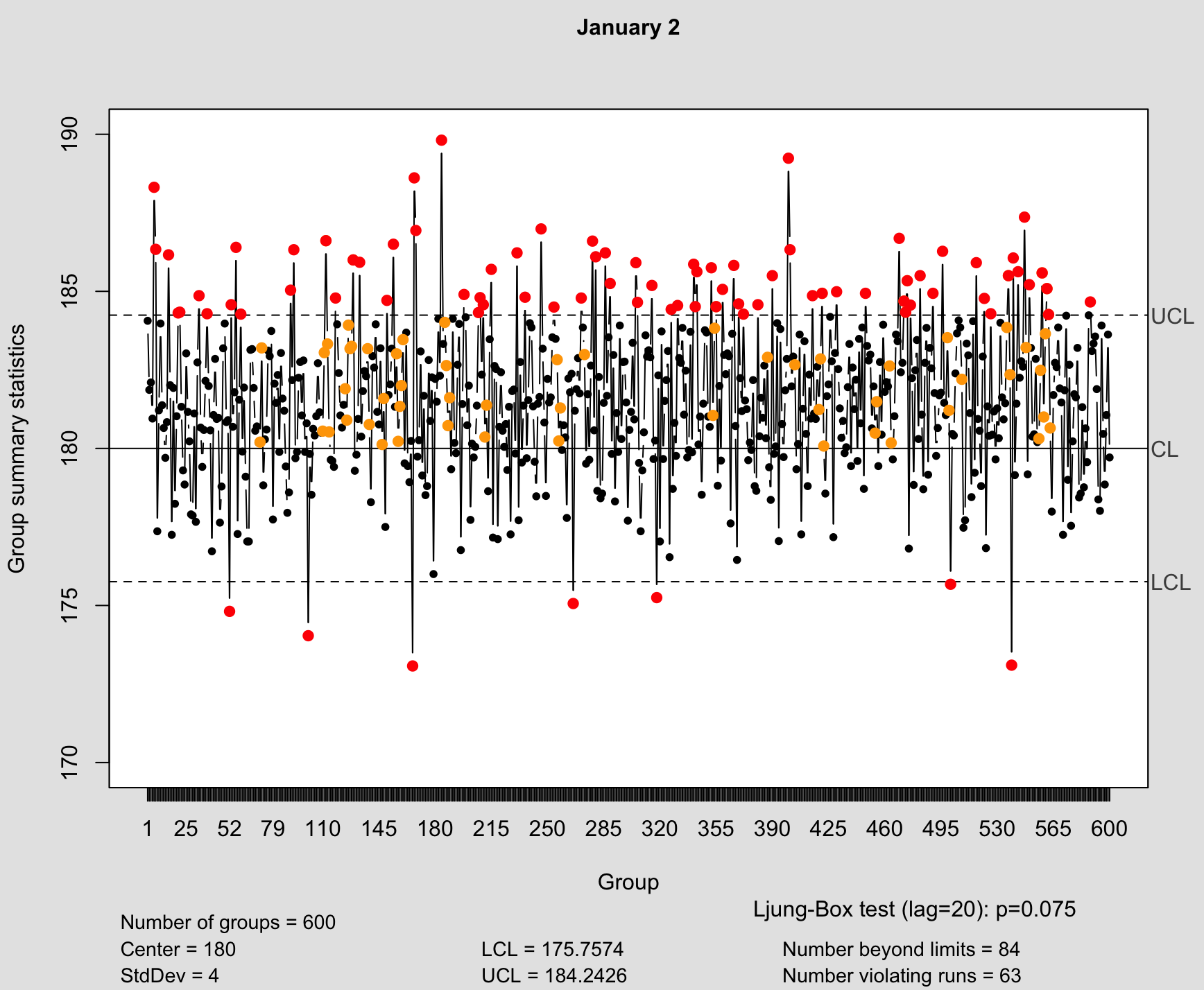}
\end{minipage}\hfill
\begin{minipage}{0.48\textwidth}
\centering
  \includegraphics[width=\linewidth]{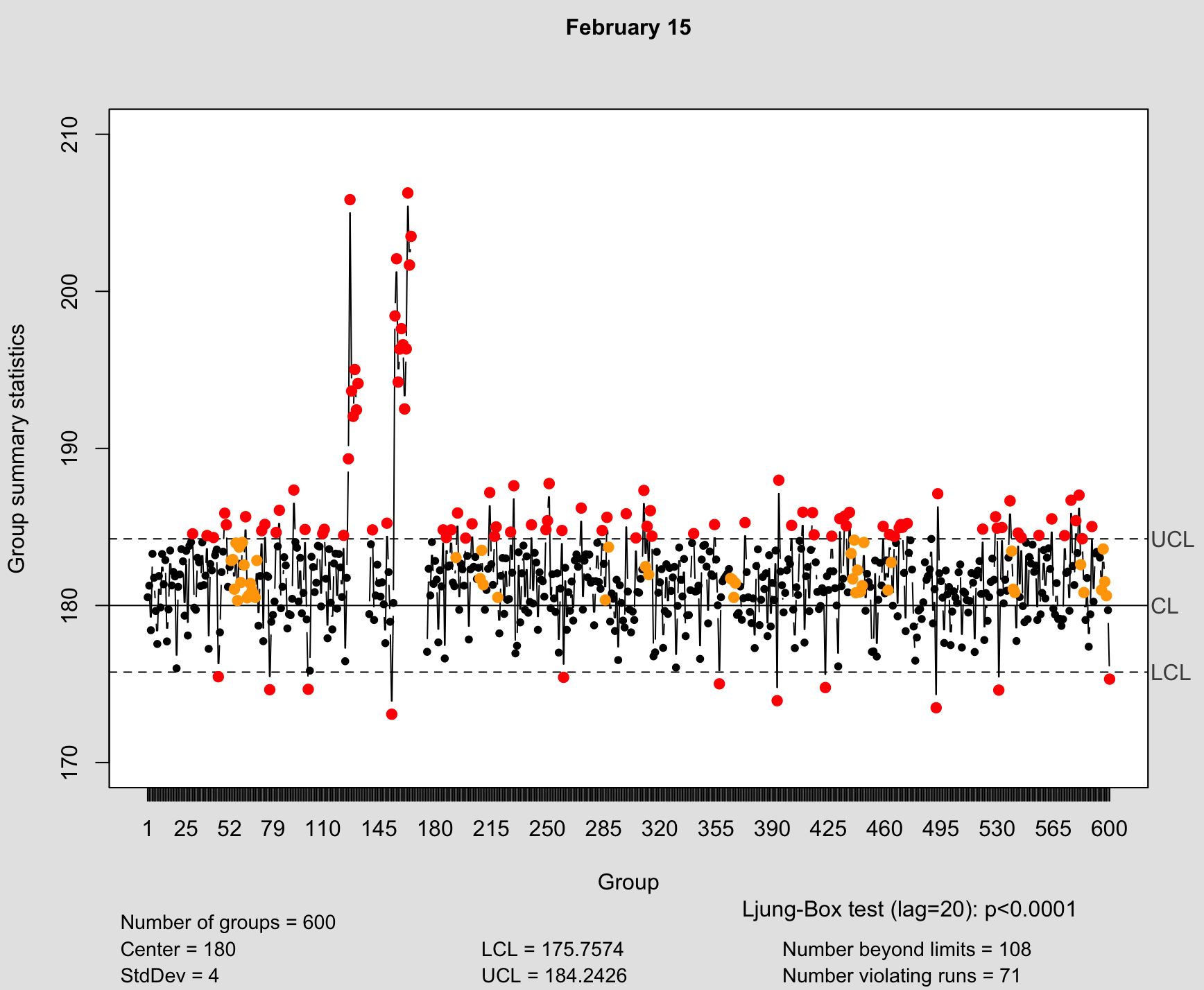}
\end{minipage}

\caption{Shewhart $\bar x$ charts for temperatures collected on January 2 and February 15}\label{CCharts}
\end{figure}

Figure~\ref{CCharts} illustrates the limitations of the use of classical methods by presenting Shewhart $\bar{x}$ charts for January~2 and February~15, the latter of which had two production halts. The control charts were generated using the R package \texttt{qcc} (v2.7). With respect to process variability, range charts (R charts) are not necessary  because the true process standard deviation ($\sigma$) is known and  assumed stable.  The results of the Ljung-Box test used to assess the independence of the sequence of sample means were added to the $\bar{x}$ chart statistics.

For both days, the charts flag process violations under Rule 1, which signals an OutC condition when a single subgroup mean exceeds the (3$\sigma$) control limits. Moreover, these violations persist under the more conservative Rule 2, because two consecutive subgroup means fall outside the limits on the same side.

Note that under InC conditions, and assuming normality and independence of subgroup means, the probability of a single false alarm (i.e., one subgroup mean plotted outside the control limits) is approximately $0.0027$ per subgroup, whereas the probability of two consecutive false alarms is $0.0027^2 \approx 7.3 \times 10^{-6}$. Consequently, under InC conditions, the Average Run Length (ARL) for  Rule~1 is $\text{ARL} = 1/0.0027 \approx 370$ subgroups, which corresponds to approximately $12.3$ hours of continuous production (1 hour of continuous production = 30 subgroups). In contrast, under  Rule~2, the ARL increases to \mbox{$1/(7.3 \times 10^{-6}) \approx 136{,}986$} subgroups, i.e., approximately 229 days (1 day of continuous production = 600 subgroups).

 The inadequate performance of classical methods in this context motivated the development of a more robust alternative to traditional SQC methods inspired by ant algorithms and CVSS (Section~\ref{Method}). The proposed approach provides effect smoothing, reducing noise from low-quality data, while highlighting the underlying signal that reliably indicates whether the process is InC or OutC.
 
\section{In-control/out-of-control classification using digital pheromones} \label{Method}

Ant colonies are considered superorganisms, as they consist of numerous individuals that interact and behave in a coordinated manner to achieve common goals. Each ant follows simple, decentralized behavioral rules, which collectively give rise to complex group dynamics. At the colony level, these interactions result in the emergence of organized, large-scale patterns. For further discussion, see Moffett et al. \cite{Moffett}.
 
Ants use pheromones as a means of communication to coordinate collective behaviors, particularly foraging activities (see Jackson and Ratnieks \cite{Jackson}). Forager ants search for food and, upon returning to the nest, deposit a chemical trail composed primarily of pheromones, which serves as a guide for other colony members (cf.\ Morgan \cite{Morgan}). Each ant that follows a trail and successfully returns with food reinforces the pheromone concentration along that path, thereby increasing its attractiveness to other ants. In contrast, trails leading to more distant or less rewarding food sources are used less frequently. As a result, their pheromone levels gradually diminish over time, ultimately leading to their abandonment.

Inspired by the foraging behavior of ants, Dorigo~\cite{Dorigo1992} developed a metaheuristic algorithm that simulates how ants explore their environment and reinforce successful paths using pheromone trails. This method is designed to efficiently identify optimal or near-optimal solutions to complex optimization problems. 
 Comprehensive overviews are provided by Dorigo and Stützle~\cite{AntCol} and Monmarché et al. \cite{ArtAnt}. In this context, digital pheromones are used to indicate promising paths; see, for example, Garnier et al. \cite{Garnier}, Fujisawa et al. \cite{Fujisawa}, Barbosa and Petty \cite{Barbosa}, and Khanduja et al. \cite{Khanduja}. For recent developments integrating these ideas into reinforcement learning frameworks, consult \cite{Fayaz,Zhang2021,Yu2023, Sousa,Wang,Hu2025}.

\begin{figure}[htbp]
\centering
\includegraphics[width=0.43\textwidth]{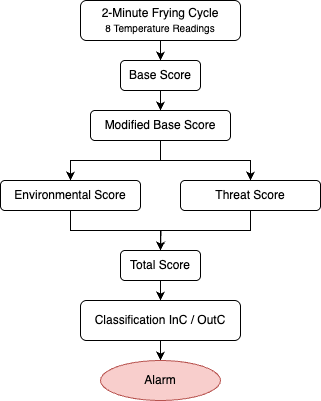}
\caption{Scoring and classification workflow}
\label{workflow}
\vspace{-1em}
\end{figure}

In adapting this approach to the potato chip frying process,  the InC/OutC classification task was addressed. Each sequence of 8 temperature readings, captured over a 2-minute frying interval, is treated as a digital ant that deposits a trail of digital pheromones. A transition from InC to OutC is signaled as a potential vulnerability in the production line. Drawing inspiration from  CVSS, a set of evaluation metrics was defined: the Base Score, Modified Base Score, Threat Score, and Environmental Score. These metrics contribute to the computation of a composite digital pheromone Total Score, which is detailed below.  To clarify the structure of the scoring and classification process, Figure~\ref{workflow} provides a schematic overview. 

\subsection{Base score}\label{BScore}

The Base Score (BS) is defined as 
\begin{equation}\label{BSeqn}
\text{BS}= \frac{\max}{\min} \cdot \frac{N_{184\uparrow}+N_{188\uparrow}+N_{192\uparrow}-N_{180\downarrow}}{2} ,
\end{equation}
where $\frac{\max}{\min}$ is the temperature max-min ratio, $N_{184\uparrow}$  the number of temperatures above 184\degree, $N_{188\uparrow}$ the number of temperatures above  188\degree, $N_{192\uparrow}$ the number of temperatures above 192\degree, and $N_{180\downarrow}$ the number of temperatures below 180\degree.

Note that in formula \eqref{BSeqn}, temperatures within the interval \mbox{$(184\degree, 188\degree]$} are counted twice, while those above 192\degree\ are counted three times. This weighting scheme amplifies the influence of higher temperature readings on the computed BS value.  The term $N_{180\downarrow}$, on the other hand, acts as a moderating factor to mitigate the impact of seemingly anomalous high temperatures in the sequence that could otherwise result in the misclassification of an InC process as OutC. As for the max-min ratio, it  captures the spread of the temperature readings,  incorporating this variability into the BS, and offering a more intuitive, scale-free measure of variability than common scale-dependent dispersion metrics such as the range or standard deviation.

If  all temperatures lie within the interval $[180\degree, 184\degree]$, indicating that the process is in an InC state, then \mbox{$\text{BS} = 0$}. 
Conversely, when all temperature readings exceed 192\degree, the multiplier  $\frac{N_{184\uparrow}+N_{188\uparrow}+N_{192\uparrow}-N_{180\downarrow}}{2}$  reaches its maximum value of 12. Given that, under an InC state, the ratio $\frac{\max}{\min} \geq 1$ and  is  unlikely to exceed \mbox{$\frac{\mu + 3\sigma}{\mu - 3\sigma} = 1 + \frac{6\sigma}{\mu - 3\sigma} \approx 1.1429$},  BS values greater than or approaching 12 provide a strong indication that the process might be operating in an OutC state.

\begin{center}
\begin{figure}[htbp]\centering
\includegraphics[height=7cm,width=10cm]{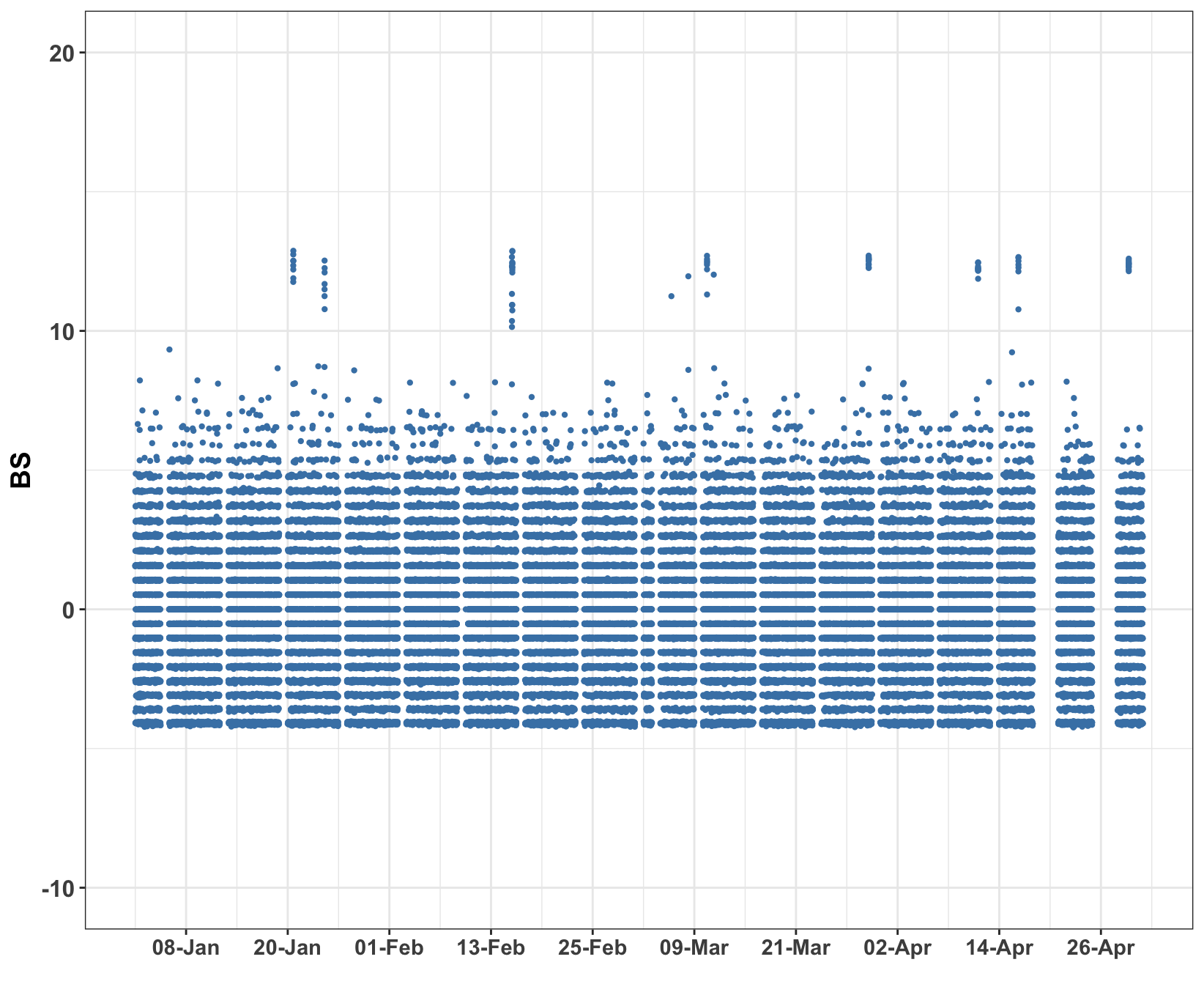}
\caption{BS values observed from January to April}\label{BSTS}
\end{figure} 
\end{center}

Figure \ref{BSTS} shows the BS values over time, highlighting 11 surges listed in Table~\ref{BSCE}. Since the extreme values recorded on 2025/03/06, 2025/03/08, and 2025/03/11 at 07:02 likely stem from lack of feedback and control during the initial phase of operation, they are disregarded, leaving 9 genuine indications to halt production.
 
\begin{table}[htbp]
\renewcommand{\arraystretch}{1.2}
\centering
\caption{BS extreme values}\label{BSCE}
\begin{threeparttable}
\begin{tabular}{|l|l@{\hspace{1.5em}}|c|} \hline
Date & Time & BS \\ \hline
2025/01/20 & 15:52 & 12.88 \\ \hline
2025/01/24 & 08:22 & 12.52 \\ \hline
2025/02/15 & 12:28 & 12.87 \\ \hline
2025/03/06 & 07:02\tnote{(1)} & 11.25 \\ \hline
2025/03/08 & 07:02\tnote{(1)} & 11.96 \\ \hline
2025/03/10 & 11:56 & 12.70 \\ \hline
2025/03/11 & 07:02\tnote{(1)} & 12.02 \\ \hline
2025/03/29 & 14:00 & 12.70 \\ \hline
2025/04/11 & 13:06 & 12.46 \\ \hline
2025/04/16 & 07:22 & 12.65 \\ \hline
2025/04/29 & 07:56 & 12.59 \\ \hline
\end{tabular}
\begin{tablenotes}
\footnotesize
\item[(1)] Discard, as it likely results from delayed feedback and control during startup.
\end{tablenotes}
\end{threeparttable}
\end{table}

\subsection{Modified base score}

The Modified Base Score (MBS) is defined as
\begin{equation*}
\text{MBS} = T_1 \times T_2 \times \text{BS} ,
\end{equation*}
where $T_1$ and $T_2$ are tuning parameters.

To compute the tuning  parameter $T_1$ for ant $k$, the BS values of the next 5 ants, $\text{BS}_{k+1}, \dotsc, \text{BS}_{k+5}$, are considered. Each value is compared with $\text{BS}_k$, and scores are assigned as follows:
\begin{itemize}
\item Assign a score of $0.1$ for each $\text{BS}_{k+j} > \text{BS}_k$;
\item Assign a score of $-0.05$ for each $\text{BS}_{k+j} < 0$.
\end{itemize}

The tuning parameter $T_1$ for ant $k$ is then calculated as
$$
T_1 = 1 + 0.1 \times \#\{\text{BS}_{k+j} > \text{BS}_k\} - 0.05 \times \#\{\text{BS}_{k+j} < 0\} ,
$$
where $j = 1, \dotsc, 5$.

The second tuning parameter, $T_2$, is defined  for ant $k$ as
$$
T_2 = \left\{
\begin{array}{ll}
1.1 & \;\; \text{if there is an increasing pattern in the last 4 temperatures} \\
0.9 & \;\; \text{if there is a decreasing pattern in the last 3 temperatures} \\
1 & \;\; \text{otherwise}
\end{array} 
\right.
$$

It is important to note that the MBS operates as a dynamic extension of the BS, incorporating context-sensitive tuning parameters to modulate pheromone levels in an adaptive manner. This mechanism enhances the responsiveness of the system by amplifying signals indicative of legitimate process-stopping conditions while attenuating those arising from spurious fluctuations or noise.

\begin{figure}[htbp]\centering
\includegraphics[height=7cm,width=10cm]{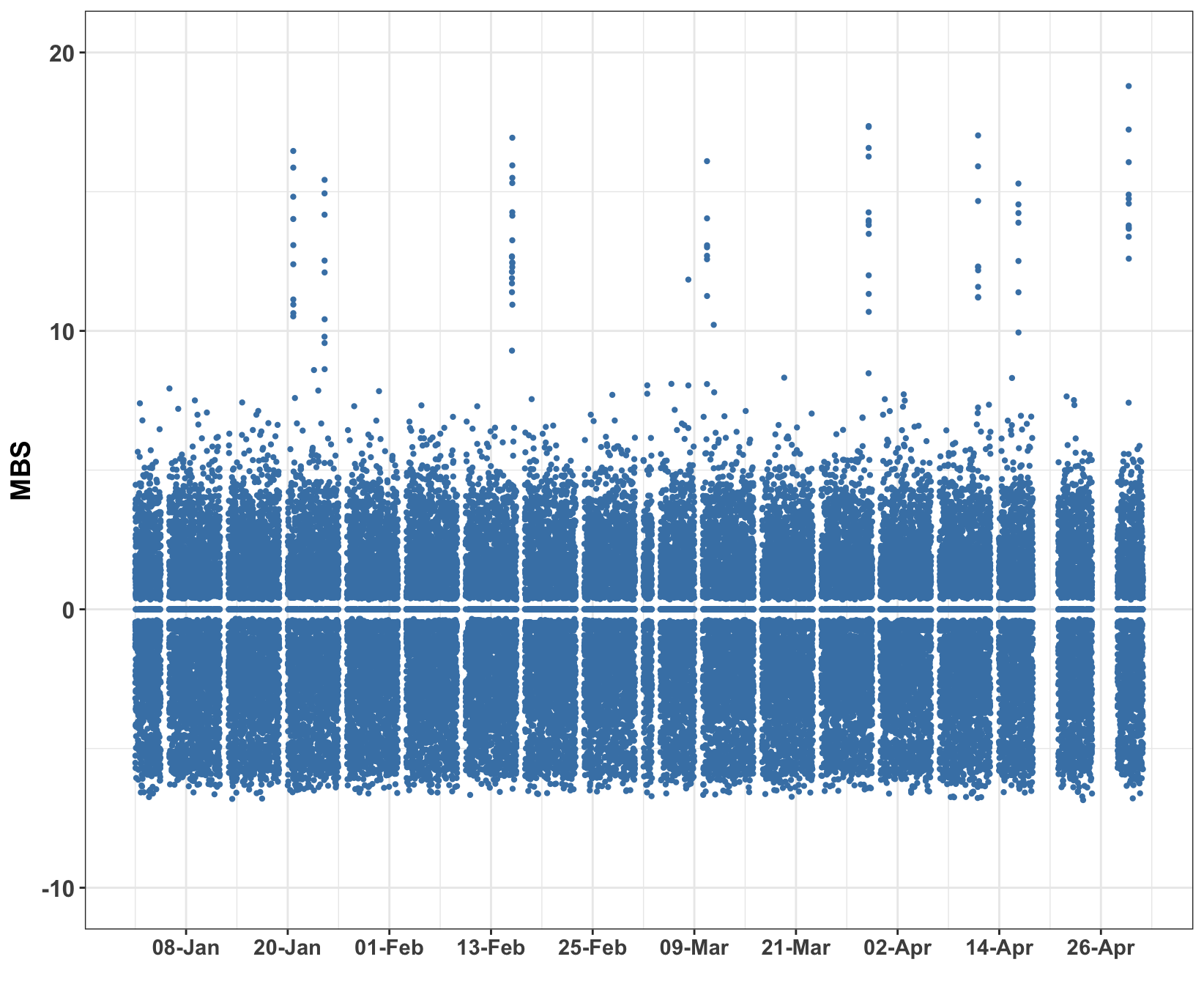}
\caption{MBS values observed from January to April}\label{MBSTS}
\end{figure}

Figure~\ref{MBSTS} displays the MBS values over time. As noted earlier, the BS values indicate 9 distinct surges. However, the MBS surges observed on 2025/03/08 and 2025/03/11, listed in Table~\ref{MBSCE},  are deemed irrelevant due to inadequate feedback and control mechanisms during the system's initial operational phase.

\begin{table}[htbp]
\renewcommand{\arraystretch}{1.2}
\centering 
\caption{MBS extreme values}\label{MBSCE}
\begin{threeparttable}
\begin{tabular}{|l|l@{\hspace{1.5em}}|c|} \hline
Date & Time & MBS \\ \hline
2025/01/20 & 15:40 & 16.46 \\ \hline
2025/01/24 & 08:24 & 15.42 \\ \hline
2025/02/15 & 12:18 & 16.94 \\ \hline
2025/03/08 & 07:02\tnote{(1)} & 11.84 \\ \hline
2025/03/10 & 11:54 & 16.10 \\ \hline
2025/03/11 & 07:02\tnote{(1)} & 10.22 \\ \hline
2025/03/29 & 13:58 & 17.36 \\ \hline
2025/04/11 & 12:56 & 17.02 \\ \hline
2025/04/16 & 07:12 & 15.29 \\ \hline
2025/04/29 & 07:42 & 18.79 \\ \hline
\end{tabular}
\begin{tablenotes}
\footnotesize
\item[(1)] Discard, as it likely results from delayed feedback and control during startup.
\end{tablenotes}
\end{threeparttable}
\end{table}

\subsection{Threat score}

The Threat Score (ThS) is defined as 
\begin{equation*}
\text{ThS} = \text{CP}+ \text{M}+\text{m}+\text{R} \,,
\end{equation*}
where  CP  represents the total number of change-point detections,  and
\[
\text{M} = \begin{cases}
1 & \text{if } \max \geq 195\degree
\\
0 & \text{otherwise}
\end{cases}
\, ,
\; 
\text{m} = \left\{ \begin{array}{cl}
-0.5 &  \text{ if   } \min \leq 174\degree
\\
0 & \text{ otherwise}
\end{array} \right.
\;
\text{and} 
\;
\text{R} = \begin{cases}
1 & \text{if  range $\geq 13\degree$}
\\
0 & \text{otherwise}
\end{cases} \, .
\]

\begin{center}
\begin{figure}[htbp]\centering
\includegraphics[height=7cm,width=10cm]{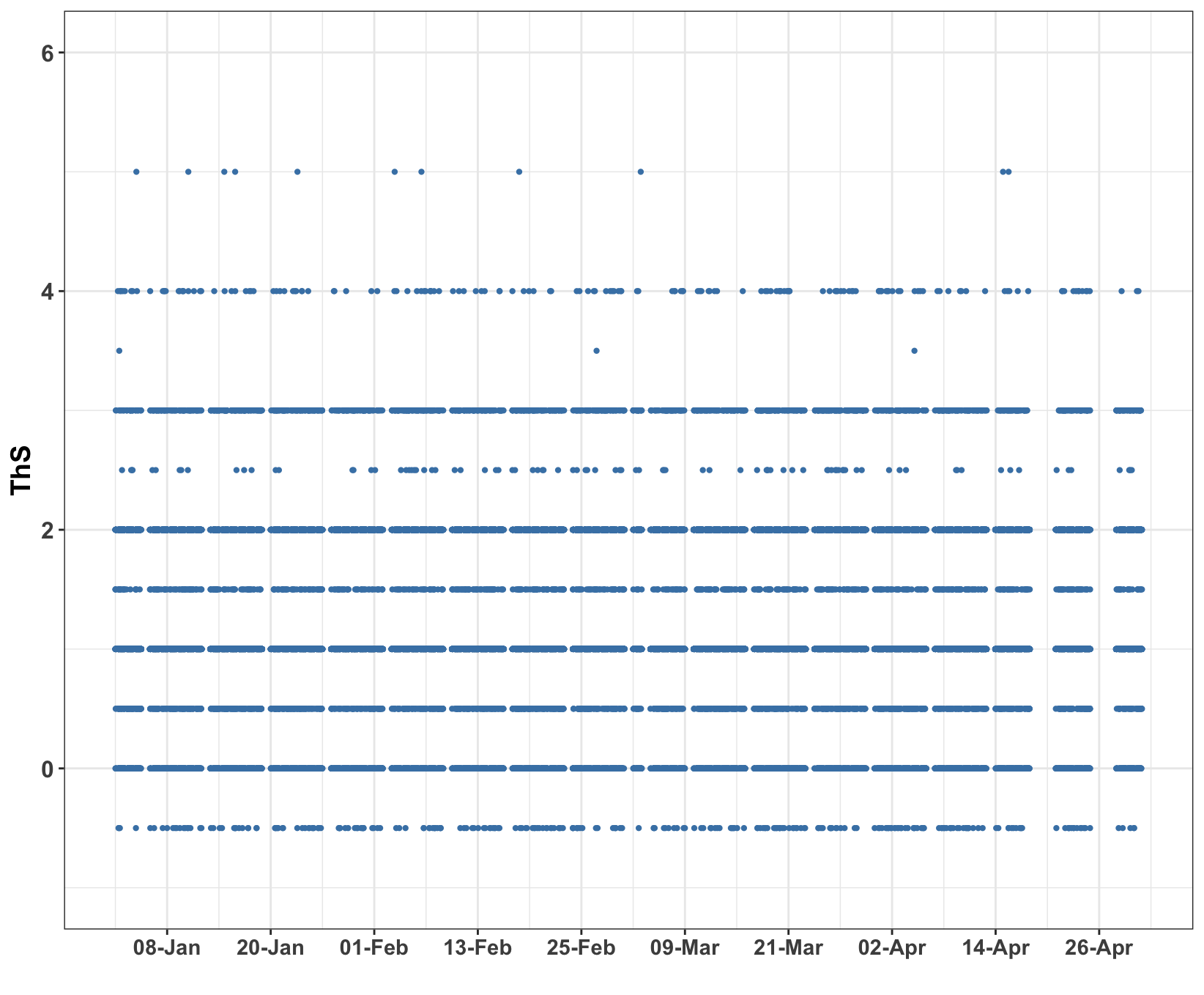}
\caption{ThS values observed from January to April}\label{ThSc}
\end{figure} 
\end{center}

Figure~\ref{ThSc} presents the ThS values over time, highlighting 11 high scores equal to 5. These high values occurred on the following dates and times: January 4 (10:24), 10 (10:54), 14 (14:56), 15 (21:06), and 23 (02:22); February 3 (08:46), 6 (11:10), and 17 (19:06); March 3 (21:02); and April 14 (21:36) and 15 (13:06). Notably, none of these high ThS occurrences align with the dates of extreme values observed in the other scores.

\subsection{Environmental score}
The Environmental Score (ES) is defined as a sum of weighted averages of the MBS values from the previous hour (i.e., for the last 30 ``ants''), incorporating the  decay of digital pheromones over time. It is computed as
\begin{equation*}
\text{ES} = \overline{\text{BS}}_{30} + \overline{\text{BS}}_{20} + \overline{\text{BS}}_{10} \, ,
\end{equation*}
where
\[
\overline{\text{BS}}_{30} = \frac{1}{2} \cdot \frac{1}{10} \sum_{k=-60}^{-42} \text{MBS}_k, \quad 
\overline{\text{BS}}_{20} = \frac{3}{4} \cdot \frac{1}{10} \sum_{k=-40}^{-22} \text{MBS}_k, \quad
\overline{\text{BS}}_{10} = \frac{1}{10} \sum_{k=-20}^{-2} \text{MBS}_k \, .
\]

\begin{center}
\begin{figure}[htbp]\centering
\includegraphics[height=7cm,width=10cm]{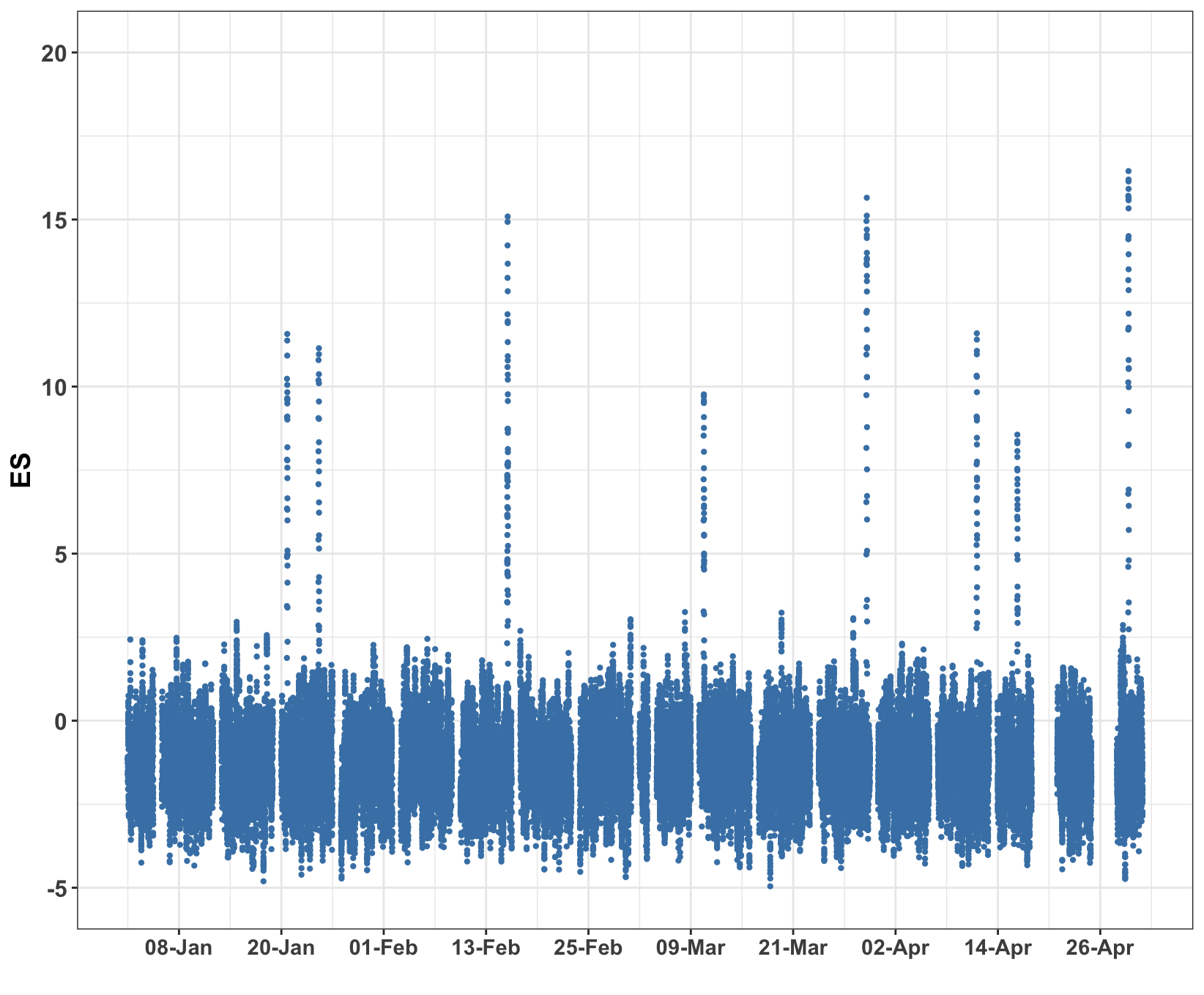}
\caption{ES values observed from January to April}\label{EnvirSc}
\end{figure} 
\end{center}

 Note that the above subscripts indicate the time windows from which the  MBS values are to be taken: 42-60 minutes prior for $\overline{\text{BS}}_{30}$, 22-40 minutes prior for $\overline{\text{BS}}_{20}$, and 2-20 minutes prior for $\overline{\text{BS}}_{10}$. The ES values over time are shown in Figure~\ref{EnvirSc}, while the extreme ES values are listed in Table~\ref{ESCE}.

\begin{table}[htbp]
\renewcommand{\arraystretch}{1.2}
  \centering
\caption{ES extreme values}\label{ESCE}
\begin{tabular}{|l|l|c|} \hline
Date & Time   & ES  \\ \hline
 2025/01/20 & 16:34 & 11.57  \\ \hline
2025/01/24 & 09:36& 11.15    \\ \hline
2025/02/15 & 12:48 & 15.09 \\ \hline
2025/03/10 & 12:42 & 9.77 \\ \hline
2025/03/29& 15:04 & 15.65 \\\hline
2025/04/11 & 13:44& 11.59\\\hline
2025/04/16 & 07:38& 8.56 \\\hline
2025/04/29 & 08:26 & 16.45 \\\hline
\end{tabular}
\end{table}

\subsection{Total score}

The Total Score (TS) is defined as
\begin{equation*}
\text{TS} = \text{MBS} + \text{ThS} + \text{ES} \, .
\end{equation*}

\begin{figure}[htbp]
\centering
\includegraphics[height=7cm,width=10cm]{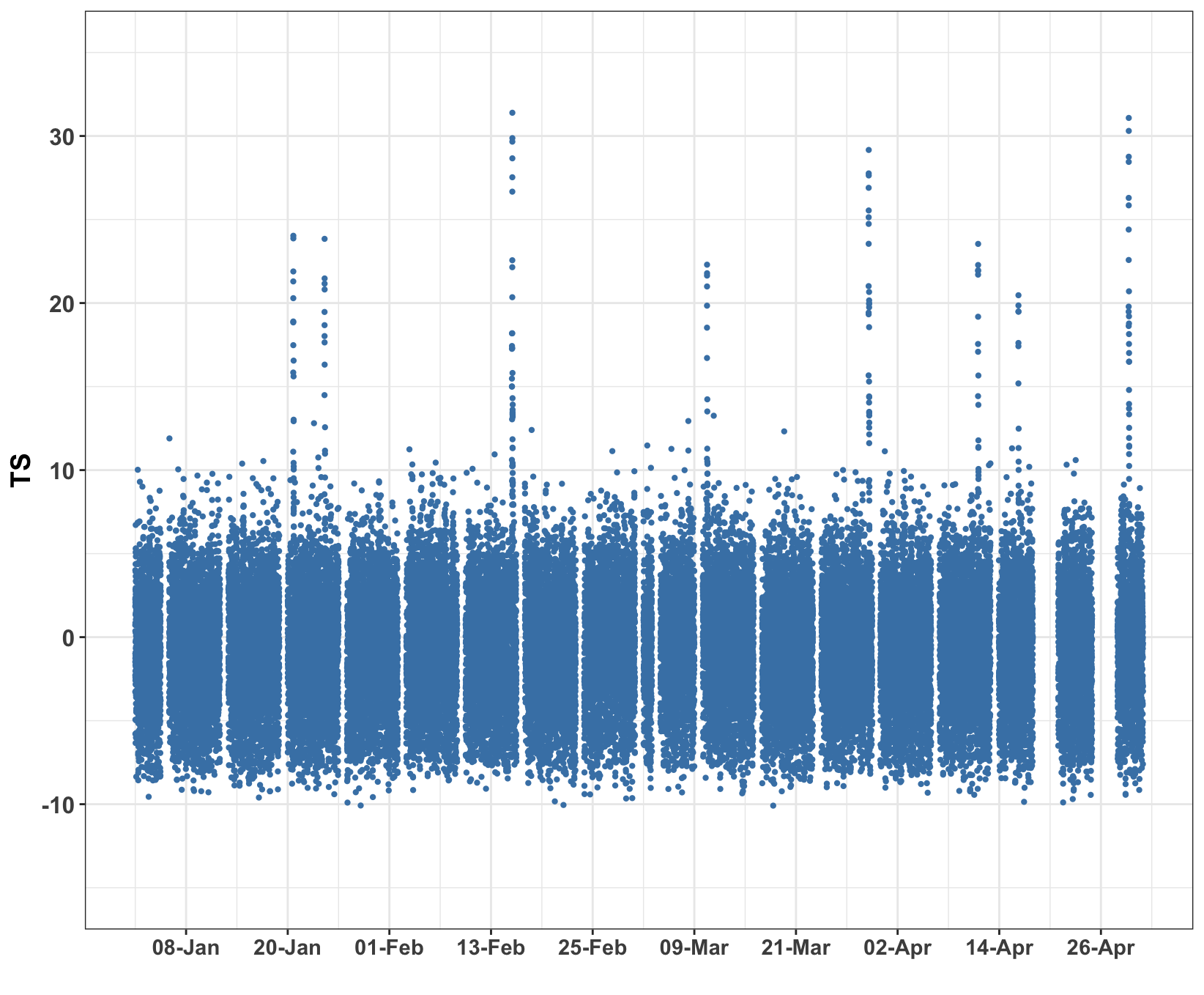}
\caption{TS values observed from January to April}\label{TotalSc}
\end{figure}

\begin{figure}[htbp]
\centering
\includegraphics[scale=0.18]{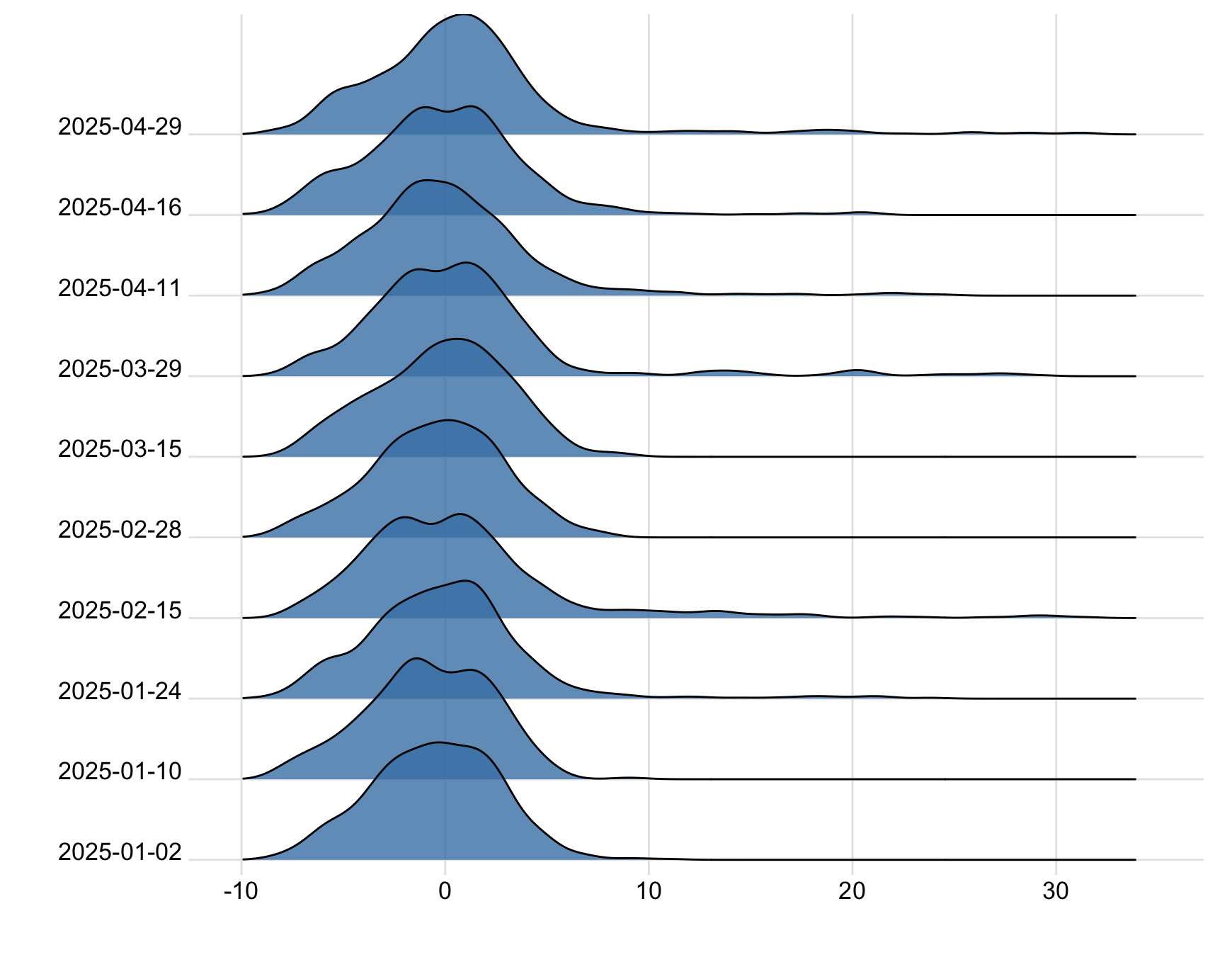}
\caption{Daily TS value distribution}\label{ridge}
\end{figure}

\begin{figure}[htbp]
\centering
\includegraphics[scale=0.2]{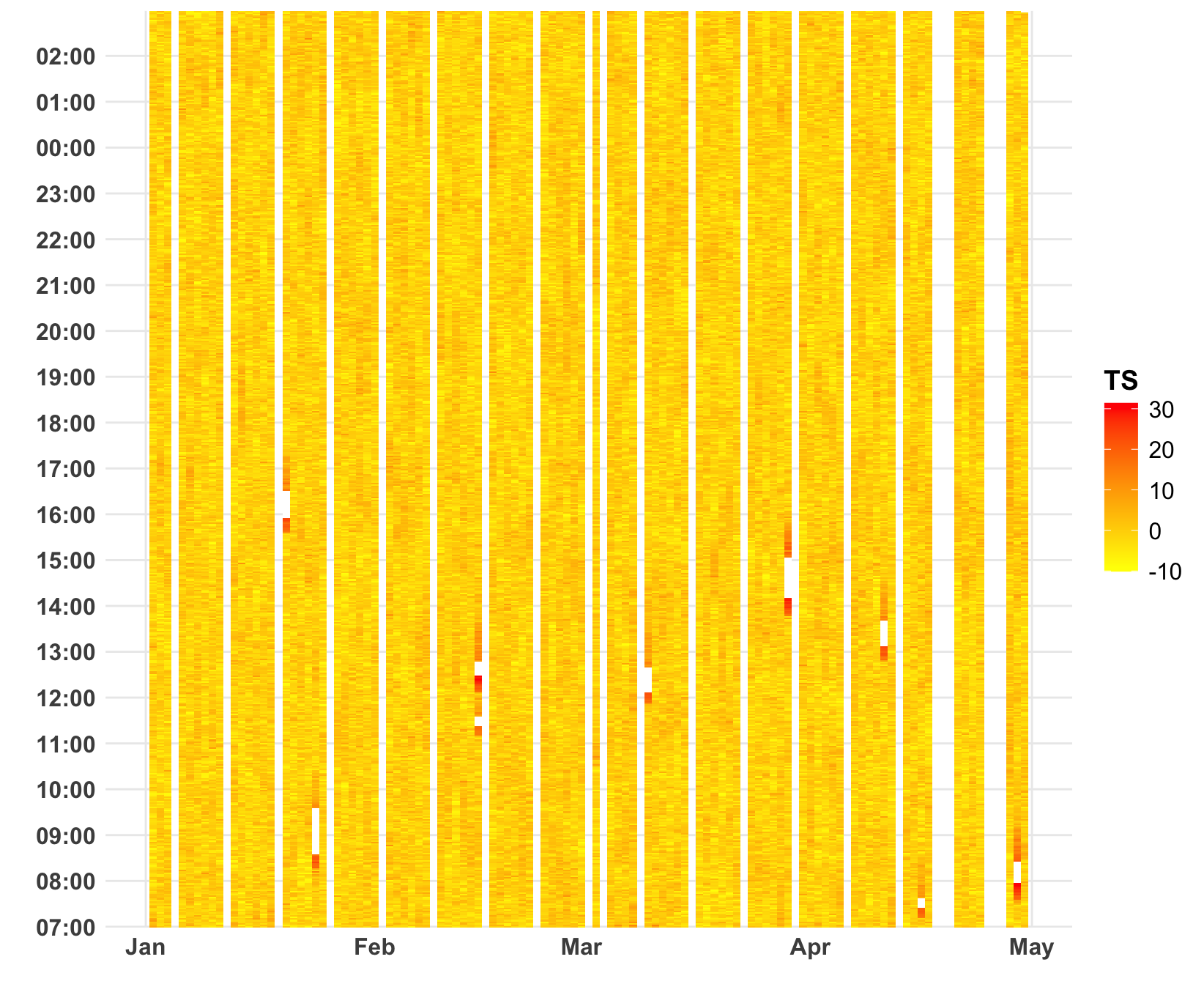}
\caption{TS value heatmap}\label{heat}
\end{figure}

Figure~\ref{TotalSc} shows the TS values over time, with Table~\ref{TSCE} summarizing 8 suspect spikes marked by extreme values. 
Figure~\ref{ridge} presents the estimated daily TS value distribution for selected dates, while Figure~\ref{heat} displays the TS value heatmap.

\begin{table}[htbp]
\renewcommand{\arraystretch}{1.2}
\centering
\caption{TS extreme values}\label{TSCE}
\begin{tabular}{|l|l|c|} \hline
Date & Time   &TS  \\ \hline
2025/01/20 & 15:52 & 24.03  \\ \hline
2025/01/24 & 08:24 & 23.84    \\ \hline
2025/02/15 & 12:24 & 31.39 \\ \hline
2025/03/10 & 12:06 & 22.30\\ \hline
2025/03/29& 14:10 & 29.16 \\\hline
2025/04/11 & 13:06& 23.54\\\hline
2025/04/16 & 07:18& 20.46 \\\hline
2025/04/29 & 07:56 & 31.08 \\\hline
\end{tabular}

\end{table}

\section{DInC/DOutC classification using digital pheromones}  \label{sec4}

\begin{figure}[htbp]
\centering
\includegraphics[scale=0.6]{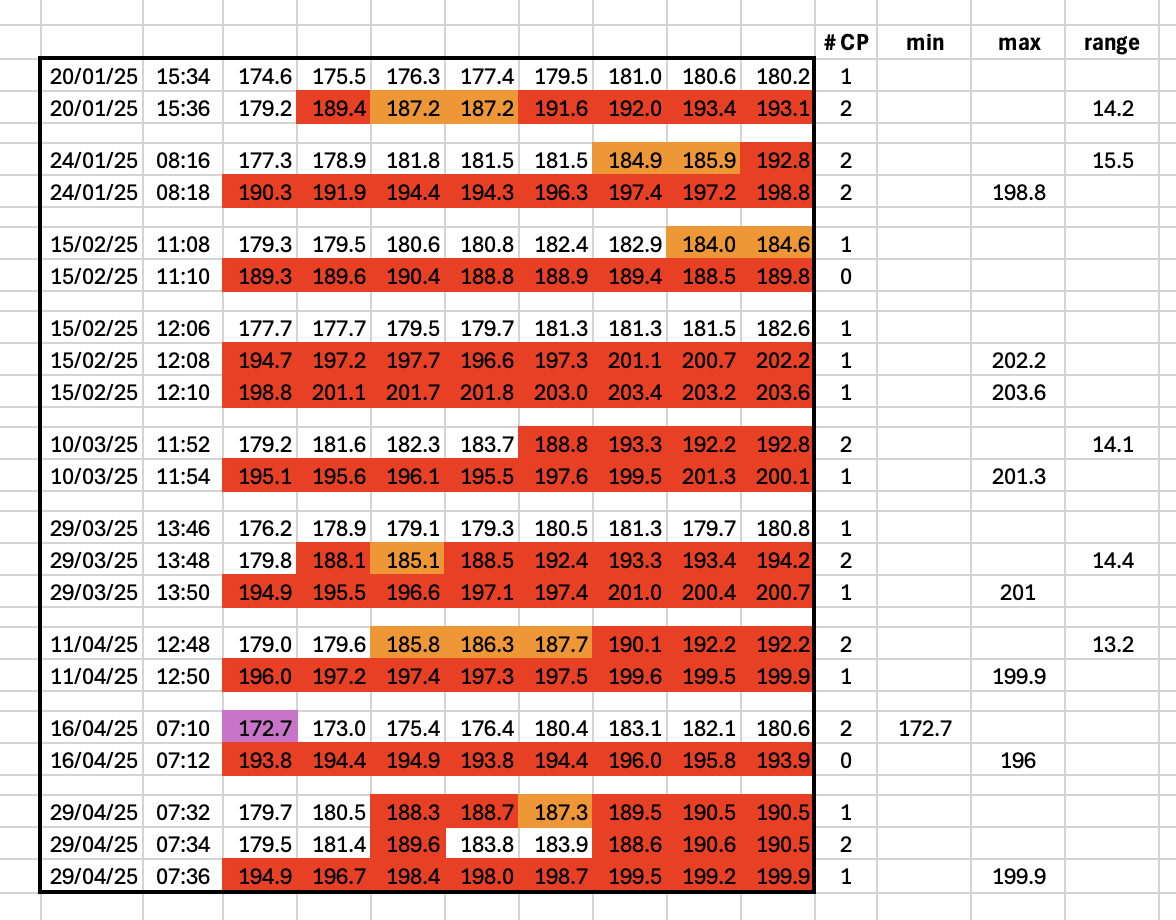}
\caption{Production interruptions between January and April}\label{Out2025}
\end{figure}

Based on the digital pheromone scores, sound and vibration alarms were activated at the dates and times listed in Table~\ref{alarm}. There were 10 production interruptions due to maintenance or repair between January and April, as identified by the InC/OutC classification algorithm and shown in Figure~\ref{Out2025}. 

\begin{table}[htbp]
\renewcommand{\arraystretch}{1.2}
\centering
\caption{Alarm activation dates and times}\label{alarm}
\begin{threeparttable}
\begin{tabular}{|l@{\hspace{1.5em}}|c|} \hline
Date & Time \\ \hline
2025/01/20\tnote{(1)} & 15:34 \\ \hline
2025/01/23\tnote{(1,2)} & 14:32 \\ \hline
2025/01/24\tnote{(1)} & 08:16 \\ \hline
2025/02/15\tnote{(1)} & 11:08 \\ \hline
2025/02/15\tnote{(1,3)} & 12:06 \\ \hline
2025/03/10\tnote{(4)} & 11:52 \\ \hline
2025/03/29\tnote{(4)} & 13:46 \\ \hline
2025/04/11\tnote{(4)} & 12:48 \\ \hline
2025/04/16\tnote{(4)} & 07:10 \\ \hline
2025/04/29\tnote{(4)} & 07:32 \\ \hline
\end{tabular}
\begin{tablenotes}
\footnotesize
\item[(1)] Training.
\item[(2)] False Discovery.
\item[(3)] False Omission.
\item[(4)] Testing.
\end{tablenotes}
\end{threeparttable}
\end{table}

The complementary contributions of BS, MBS, ES, and TS yield the confusion matrix and related performance metrics presented in Table~\ref{MarApr}. This result is considered acceptable, given the high values of both sensitivity and specificity, and the fact that the trade-off between them appropriately emphasizes sensitivity.

\begin{table}[htbp] 
\centering
\small
\setlength{\tabcolsep}{3.5pt} 
\caption{Confusion matrix (March–April)}\label{MarApr}
\begin{tabular}{|c|c|c|c|c|} \hline
Total = 14 & DInC & DOutC & BM = 0.55 & PT = 0.3586 \\ \hline
OutC (P) = 10 & TP = 8  & FN = 2  & TPR = 0.8 & FNR = 0.2 \\ \hline
InC (N) = 4   & FP = 1  & TN = 3  & FPR = 0.25 & TNR = 0.75 \\ \hline
Prev = 0.7143 & PPV = 0.8889 & NPV = 0.6 & LR${}^+$ = 3.2 & LR${}^-$ = 0.2667 \\ \hline
Acc = 0.7857  & FDR = 0.1111 & FOR = 0.4 & MK $\Delta_P$ = 0.4889 & DOR = 12 \\ \hline
BAcc = 0.775  & $F_1$ Score = 0.8421 & FM Index = 0.8433 & MCC = 0.5185 & CSI = 0.7273 \\ \hline
\end{tabular}
\end{table}

It is important to emphasize that the data exhibit considerable noise, including random fluctuations and potential errors, which obscure genuine patterns and insights. Properly accounting for this noise is critical to reliable data analysis and model construction, as it can otherwise compromise the validity of the results. Consequently, the observed limitations in forecasting accuracy are not surprising.

\section{Conclusions}\label{Conclusions}

The factory’s primary objective was to predict imminent transitions from InC to OutC, and our initial intervention, therefore, focused on advising how better data could be obtained for this task, given the low-quality records observed in 2024 when we were first consulted. However, the data collected between January and April 2025 remained messy. In this context, our digital ant-pheromone-inspired methodology proved effective for InC/OutC classification, but it could not elicit any reliable, actionable rule for forecasting when a halt for maintenance or repair would be needed. This is a central limitation of the present study because classification performance does not automatically translate into forecasting.

Operational observations further illustrate this challenge. The rheostat controlling the electric current can rapidly adjust frying temperatures; even when temperatures exceed 195\degree, it appears capable of returning to the target range. Yet production interruptions on January 24, February 15 (twice), April 16, and April 29 followed sequences of moderately elevated temperatures that did not trigger timely corrective adjustments. At the same time, the introduction of additional anomaly-correction devices may have been excessive. An unexpectedly large number of ``ant'' sequences exhibited all 8 temperature readings below the 180\degree\ threshold, producing undercooked batches and undermining crispness, de-oiling efficiency, and seasoning quality. These mixed signals, namely, episodes of elevated temperatures without timely correction, alongside frequent low-temperature sequences consistent with overcorrection, are precisely the kind of confounding patterns that make forecasting rules fragile when the underlying data stream is noisy.

Despite these limitations, several elements were clearly valuable in practice. The supervision team reported that the real-time temperature display, enhanced with color coding for surges (both high and low), extreme values, and change-point detections, was highly effective for visually tracking process evolution. They also considered the sound and vibration alerts, activated when Base Score, Modified Base Score, Environmental Score, or Total Score indicated a possible shift towards an OutC state, to be appropriate and useful. Beyond immediate operational relevance, this case study demonstrates the promise of digital ants and digital pheromones, concepts from AC superorganism modeling, as ingredients in SQC classification algorithms, particularly, as a way to aggregate weak local signals into a global indication of process state.

It is important, however, to emphasize that this first application of digital scoring in an SQC context remains exploratory. The digital pheromone scores are informed by prior experience with markers that distinguish InC from OutC states, and this reliance on expert-informed structure is both a strength and a limitation. While the Total OutC Score generally yielded accurate classifications, the current formulation of the Threat Score  did not significantly improve decision-making. This suggests that some components of the scoring architecture may require reformulation and that the apparent signal may, in part, reflect artifacts of the data collection process rather than stable precursors of true OutC events.

Finally, the environmental context matters for both interpretation and evaluation. Based on 2024 data, the estimated probability of OutC surges is relatively low (approximately 0.008), and in most cases feedback and control mechanisms are sufficient to restore the InC state. Consequently, only the accumulation of clustered ``suspicious ants'', which reinforce digital pheromones and thereby increase the Environmental Score, can reliably signal an OutC event. This low base rate also means that forecasting rules must be assessed with particular care: even small increases in false alarms can rapidly erode trust and usefulness, while overly conservative thresholds may miss the rare but consequential events.

Future work should therefore prioritize disciplined data acquisition so that candidate forecasting rules can be learned, tested out-of-sample, and evaluated under realistic sensitivity-specificity requirements. Only with higher-quality, well-structured data will it be possible to determine whether an adequate and stable forecasting rule exists for anticipating imminent OutC transitions and the associated need to halt production for maintenance or repair.

\end{document}